\documentclass[conference]{IEEEtran}

\usepackage[T1]{fontenc}
\usepackage{amsmath}
\usepackage{amssymb}
\usepackage{graphicx}
\usepackage{booktabs}
\usepackage{algorithm}
\usepackage{algorithmic}
\usepackage{xcolor}
\usepackage{caption}
\usepackage{subcaption}
\usepackage{url}
\usepackage{microtype}
\usepackage{lipsum}

\usepackage{cuted}
\usepackage[hidelinks,breaklinks=true]{hyperref}
\usepackage[capitalise,noabbrev]{cleveref}

\newcommand{\Strat}{\mathcal{S}}   
\newcommand{\cl}{\operatorname{cl}}  

\newtheorem{definition}{Definition}

\begin{document}

\title{Contact Modes Are Strata:\\What Geometric Structure Buys in Discrete--Continuous Planning}

\author{%
  \IEEEauthorblockN{Phone Thiha Kyaw and Jonathan Kelly}
  \IEEEauthorblockA{Space and Terrestrial Autonomous Robotic Systems (STARS)
    Laboratory\\
    University of Toronto Institute for Aerospace Studies (UTIAS)\\
    Toronto, Ontario M3H 5T6, Canada\\
    \texttt{\{phone.thiha,jonathan.kelly\}@robotics.utias.utoronto.ca}}
}

\maketitle

\begin{strip}
\centering
\vspace{-3.5em}
\includegraphics[width=\textwidth]{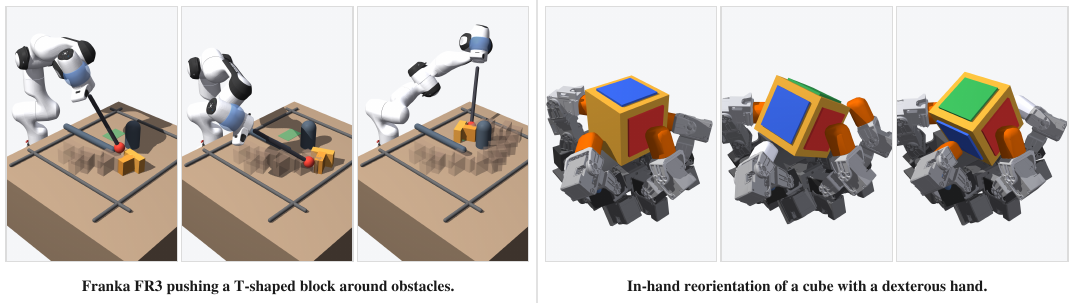}
\vspace{-1.5em}
\captionof{figure}{
Two contact-rich manipulation tasks, each planned as a walk over strata, with frames running left to right along one plan.
We give neither plan a mode, a contact sequence, or a stratum in advance.
The contacts a plan makes and breaks follow from the stratification of the configuration space, and we use the same planner for both tasks.
}
\vspace{-0.2em}
\label{fig:systems}
\end{strip}

\begin{abstract}
Contact-rich manipulation poses a discrete question and a continuous one at once, namely which contacts are active and how to move while they hold.
The two are coupled by a change of dimension, since each contact that a robot maintains confines its motion to a lower-dimensional manifold.
We make that coupling the explicit object of planning by observing that a contact mode is not merely analogous to
a stratum of the configuration space; it is one.
A plan is then a walk over strata whose within-stratum segments are geodesics.
On two contact-rich manipulation tasks in simulation, pushing a T-shaped block around obstacles and reorienting a cube in a dexterous hand, our planner returns solutions within seconds with no mode, contact sequence, or stratum given in advance.
\end{abstract}

\vspace{-0.4em}
\section{Introduction}
\label{sec:intro}

\looseness=-1
Contact-rich manipulation is a mixed discrete--continuous problem: a discrete choice of which contacts are active, interleaved with continuous motion while they hold.
The two are coupled by a change of dimension.
When a robot moves in free space, it may translate in every direction, but the moment it makes a contact, its motion is confined to a lower-dimensional manifold, and each additional contact confines it further.
A configuration space assembled from manifolds of several dimensions, glued along their boundaries, is a \emph{stratified space},
and the manifolds from which the space is assembled are its \emph{strata} (Definition~\ref{def:stratified}).

This observation is not new.
Stratified configuration spaces were studied early on and entered robotics through legged locomotion and finger gaiting~\cite{goodwine2002motion,goodwine1998thesis}, later finding application in stance stability~\cite{rimon2008general}, closed kinematic chains~\cite{han2008convexly}, and quadruped motion planning~\cite{bhattacharya2007stratified}.
Bretl et al. then set the representation aside as ``impractical for free-climbing robots'' in~\cite{bretl2005multi}.
This was true in 2005, but since then, continuation methods on implicitly defined constraint manifolds have matured~\cite{jaillet2012path,kingston2019exploring,kingston2018sampling}, and collision engines have become much faster~\cite{thomason2024motions,sundaralingam2023curobo}.

\looseness=-1
Constrained motion planning methods often assume a single manifold of fixed dimension, given in advance in the form of implicitly defined equality constraints.
A stratified configuration space has no such single manifold.
It has one manifold, of its own dimension, for every set of contacts that can hold simultaneously, and choosing among these sets is the discrete part of the problem.
In this work, we show that this partition into strata of varying dimension lets the discrete choice emerge from planning itself, so no mode or contact sequence needs to be fixed in advance.
Existing frameworks for discrete--continuous planning, such as Graph of Convex Sets~\cite{morozov2025mixed}, Logic-Geometric Programming~\cite{toussaint2015lgp}, and multi-modal motion planning~\cite{hauser2010multi}, all require the discrete structure to be given explicitly, as a convex decomposition, a symbolic domain, or a mode family, and one line of work even takes the manifold sequence itself as input~\cite{englert2020sampling}.
In this work, the discrete structure is not specified a priori; instead, the strata and the transitions between them follow directly from the equality constraints a planner already evaluates.

We present the following contributions.
\begin{itemize}
  \looseness=-1
  \item We provide a formulation of stratified configuration spaces for discrete--continuous systems, and construct the stratification from only the signed distances and their gradients.
  \item We formulate discrete--continuous planning as planning over a stratified configuration space, in which a discrete mode is a stratum and a plan is a walk over strata.
  \item We report preliminary simulation results on two challenging contact-rich manipulation tasks, which the planner solves within seconds on every trial, with no mode, contact sequence, or stratum given in advance.
\end{itemize}

\section{Stratified Configuration Spaces}
\label{sec:formalism}

\looseness=-1
This section makes the link between contact modes and strata precise.
Let $q \in \mathbb{R}^n$ hold the coordinates of the robot and of the objects it manipulates,
and let $\mathcal{I}$ be the fixed set of body pairs that can come into contact.
For each pair $i \in \mathcal{I}$, the \emph{signed distance} $\varphi_i(q)$ is positive when the two bodies are apart, zero when they touch, and negative when they overlap.
The collision-free set is $\mathcal{Q}_{\mathrm{free}} = \{\, q \mid \varphi_i(q) \geq 0 \;\; \forall i \in \mathcal{I} \,\}$,
and the contacts that hold at $q$ form the \emph{active set} $A(q) = \{\, i \in \mathcal{I} \mid \varphi_i(q) = 0 \,\}$.
A collision engine reports $\varphi_i$ and $\nabla \varphi_i$ for any pair we ask about, and the construction below needs nothing more.

\looseness=-1
We evaluate the active set at each configuration $q$ visited during planning rather than choose it in advance, and it identifies the stratum containing $q$.
For each fixed $A \subseteq \mathcal{I}$, the configurations at which exactly the contacts in $A$ hold form the set
\begin{equation}
\label{eq:stratum}
\Strat_A = \{\, q \mid \varphi_i(q) = 0 \;\; \forall i \in A, \;\;
\varphi_j(q) > 0 \;\; \forall j \notin A \,\}.
\end{equation}
Every configuration has one active set, so the sets $\Strat_A$ are disjoint and together cover $\mathcal{Q}_{\mathrm{free}}$.
The strict inequality on the inactive pairs keeps the sets disjoint; a non-strict one would let $\Strat_A$ absorb every configuration that holds more contacts than $A$.

\begin{figure*}[!t]
\centering
\vspace{-1em}
\includegraphics[width=\textwidth]{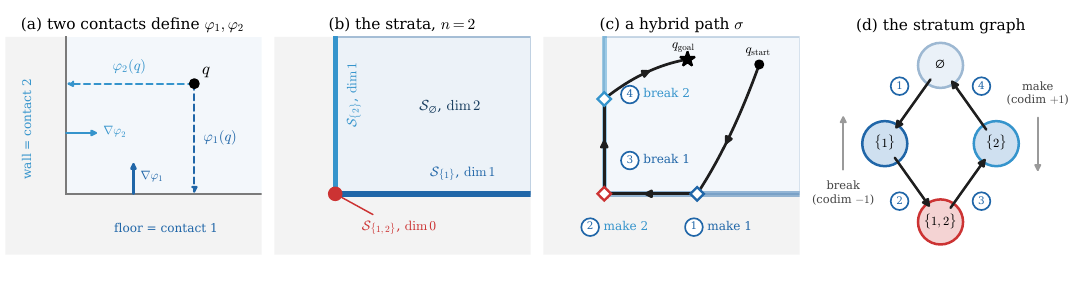}
\vspace{-3.5em}
\caption{
Contacts stratify the configuration space of a point robot between a floor and a wall ($n = 2$).
(a) Each signed distance gives the gap to one surface and vanishes where the robot touches it.
(b) Grouping configurations by which distances vanish gives four strata, of dimension $n - |A|$.
(c) A hybrid path $\sigma$ lies within strata and crosses between them where one contact is made or broken (diamonds).
(d) The same crossings on the stratum graph, numbered as in (c), where making a contact raises the codimension by one and breaking one lowers it.
}
\vspace{-1.2em}
\label{fig:strat}
\end{figure*}

\looseness=-1
We collect the active constraints into the stacked map $h_A(q) = (\varphi_i(q))_{i \in A}$, whose Jacobian $J_A(q) \in \mathbb{R}^{|A| \times n}$ has the row $\nabla \varphi_i(q)^\top$ of each active contact.
We assume throughout that these rows are linearly independent at every $q \in \Strat_A$, the \emph{linear independence constraint qualification}, so $J_A(q)$ has full row rank and $h_A$ is a submersion there.
The inactive pairs stay apart on the open set $U_A = \{\, q \mid \varphi_j(q) > 0 \;\; \forall j \notin A \,\}$, and $\Strat_A = h_A^{-1}(0) \cap U_A$ by~\eqref{eq:stratum}, so the regular value theorem makes $\Strat_A$ an embedded submanifold of $\mathbb{R}^n$ with
\begin{equation}\label{eq:dim}
  \dim \Strat_A = n - |A|, \qquad \mathcal{T}_q \Strat_A = \ker J_A(q).
\end{equation}
The dimension follows from the rank--nullity theorem, since each contact that holds contributes one independent row to $J_A(q)$ and removes one dimension from its kernel.
Free space is itself a stratum, with $A = \emptyset$, open in $\mathbb{R}^n$ and of full dimension $n$.
The equality $\varphi_i = 0$ constrains the normal gap alone, so a contact on $\Strat_A$ is free to slide along the bodies.
We show in Section~\ref{sec:cones} how holding a contact fixed on the bodies confines motion to a lower-dimensional submanifold of $\Strat_A$.

\looseness=-1
The strata are not laid out side by side but glued along their boundaries.
Along a sequence in $\Strat_A$ converging to $q$, the equalities hold in the limit by continuity and weakens the strict inequalities on the inactive pairs, so $A(q) \supseteq A$ and
\begin{equation}\label{eq:frontier}
\cl(\Strat_A) \subseteq \bigsqcup_{B \supseteq A} \Strat_B.
\end{equation}
A stratum $\Strat_B$ meeting $\cl(\Strat_A)$ therefore holds every contact of $A$, and by~\eqref{eq:dim} it has $|B| - |A|$ fewer dimensions than $\Strat_A$, one for each contact it adds.
A decomposition glued in this way is a stratified space, and we use the standard definition~\cite{goresky1988stratified}.

{\parindent0pt
\vspace{0.4em}
\begin{definition}[Stratified space]
\label{def:stratified}
A stratification of a topological space $X$ is a decomposition
$X = \bigsqcup_i \Strat_i$ into a locally finite collection of disjoint,
connected, locally closed smooth manifolds $\Strat_i$, called strata, of
possibly different dimensions, satisfying the frontier condition that
whenever $\Strat_i \cap \cl(\Strat_j) \neq \emptyset$ with $i \neq j$, the
stratum $\Strat_i$ lies in the frontier of $\Strat_j$, that is
$\Strat_i \subseteq \cl(\Strat_j) \setminus \Strat_j$.
\end{definition}
\vspace{0.2em}
}

\smallskip
\noindent\textbf{Example (planar point robot).}
Consider a point robot moving in the plane between a floor and a wall, so $n = 2$ and $\mathcal{I}$ holds two pairs (Figure~\ref{fig:strat}a).
The signed distances $\varphi_1$ and $\varphi_2$ measure the gap to each surface.
Grouping configurations by which of the two distances vanish splits $\mathcal{Q}_{\mathrm{free}}$ into four strata (Figure~\ref{fig:strat}b).
Away from both surfaces, the robot may translate in every direction, so $\Strat_{\emptyset}$ is the open interior, of dimension two.
On the floor, only sliding along it keeps the gap closed, which leaves the edge $\Strat_{\{1\}}$ of dimension one, and the wall gives $\Strat_{\{2\}}$ in the same way.
In the corner, no direction keeps both gaps closed, so $\Strat_{\{1,2\}}$ is a single point of dimension zero.
These dimensions are the count $n - |A|$ by~\eqref{eq:dim}, since every contact that holds removes one direction.
Neither edge contains the corner, since a configuration touching both surfaces violates the strict inequality of~\eqref{eq:stratum} and therefore belongs to $\Strat_{\{1,2\}}$ alone.
A robot sliding along the floor toward the corner stays in $\Strat_{\{1\}}$ and comes arbitrarily close to it.
The corner therefore lies in the closure, and $\cl(\Strat_{\{1\}}) = \Strat_{\{1\}} \sqcup \Strat_{\{1,2\}}$ by~\eqref{eq:frontier}.
That corner is a whole stratum lying in the frontier of both edges and of the interior, as Definition~\ref{def:stratified} requires.

\subsection{The Stratum Graph}
\label{sec:graph}

\looseness=-1
The frontier relation in~\eqref{eq:frontier} makes the strata the vertices of a graph, and we join $\Strat_A$ to $\Strat_B$ when $\Strat_B$ lies in the frontier of $\Strat_A$.
By~\eqref{eq:frontier}, such a pair must satisfy $A \subsetneq B$, so crossing that edge makes the contacts in $B \setminus A$ and crossing it the other way breaks them.
We call this the \emph{stratum graph} (Figure~\ref{fig:strat}d), and our approach does not require us to construct it explicitly.
Its vertices are subsets of $\mathcal{I}$, so writing the graph down means deciding for each of the $2^{|\mathcal{I}|}$ subsets whether any configuration holds exactly those contacts.
Most such active sets are infeasible because the corresponding contacts cannot occur simultaneously.

\looseness=-1
We create an edge by motion rather than by enumeration.
We make a contact by closing the gap $\varphi_j$ while holding $h_A$ at zero, and break it by opening that gap again.
We then evaluate $A(q)$ at the resulting configuration of the motion to decide which stratum it has arrived in, rather than assuming the set it aimed for.
Because we evaluate the active set rather than assume it, we may make several contacts at once without special treatment.
We can make only the contacts whose gap at $q$ is small, so the number of edges available at $q$ grows with how many bodies are near and not with the size of $\mathcal{I}$.

Definition~\ref{def:stratified} requires the connected strata, and the sets in~\eqref{eq:stratum} need not be connected.
When no motion within $\Strat_A$ joins two ways of holding the same contacts, they lie in separate components, and the definition counts each as a stratum of its own.
We therefore use the active set as an index for a possibly disconnected union of strata; the planner does not distinguish connected components a priori.
Recovering those components would require a roadmap of $\Strat_A$, which we do not attempt here.

\subsection{Contact Cones and Foliation}
\label{sec:cones}

\looseness=-1
The robot cannot move in every direction of $\mathcal{T}_q \Strat_A$.
The coordinates of a manipulated object, a subset of the components of $q$, are \emph{passive},
meaning they are unactuated and change only through contact.
Underactuation therefore enters as constraint rows as well.
Because the manipulated object is passive, it should not move unless the active contacts can support its motion.
We therefore impose task-specific minimum-contact conditions; below that threshold, the object's coordinates are held fixed.
An object therefore moves only on the strata that meet this minimum, and the remaining strata are where the robot rearranges itself.
On those remaining strata, the extra rows confine motion to a subset of $\Strat_A$ of dimension $n - |A| - n_{\mathrm{p}}$, with $n_{\mathrm{p}}$ the number of passive coordinates, whenever the active gradients are independent of those coordinates.
Counting contacts is only a conservative approximation of the true condition, namely whether an admissible contact wrench can generate the object velocity a motion demands.

\looseness=-1
A contact also pushes and cannot pull.
At a contact through which the robot moves the object, the velocity of the contact point on the object must lie in a cone about the inward contact normal.
We call this a \emph{contact cone}, and we write it in configuration-space velocities rather than in the object twists of the pushing literature~\cite{mason1986mechanics,lynch1996stable,mason2001mechanics,chavandafle2020planar}.
Taking the contact point on the object rather than its center keeps the condition correct when the object may rotate.
The cone therefore rejects a motion that turns the object about that point, since such a motion leaves the point stationary.
The condition is one-sided and acts on velocities, so it restricts the directions available within $\Strat_A$ without changing $\dim \Strat_A$.
It approximates quasi-static pushing kinematically rather than through a limit-surface model~\cite{goyal1991planar}.

\looseness=-1
Holding a contact fixed on the bodies restricts motion differently, and unlike the cone it lowers dimension.
A contact that \emph{sticks} keeps the same point of the object in contact.
Making it records that point in the body frame of the object, and holding it replaces the single row $\varphi_i = 0$ by three rows that pin the contact to the recorded point.
Where those rows keep constant rank, they \emph{foliate} $\Strat_A$ into leaves, one per recorded point, and motion within a leaf is sticking while motion across leaves is sliding.
Each sticking contact therefore removes two further dimensions, and a leaf has dimension $n - |A| - 2|A_{\mathrm{s}}|$ with $A_{\mathrm{s}} \subseteq A$ the sticking contacts, below $\dim \Strat_A$ rather than equal to it.
A leaf is indexed by a point of a surface, so the leaves cannot be listed in advance, and two configurations that share an active set but lie on different leaves are distinct states of the search.
Reaching a distant target without sliding therefore requires breaking a contact and re-establishing it on another leaf.

\begin{figure*}[t]
\centering
\vspace{-0.5em}
\includegraphics[width=0.98\textwidth]{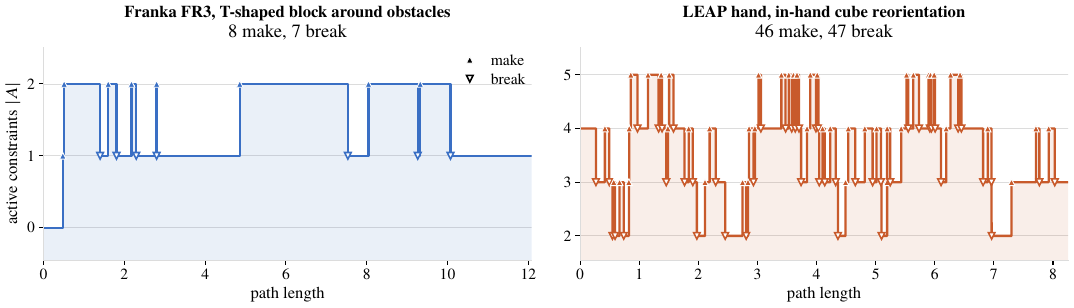}
\vspace{-0.5em}
\caption{
One plan per task, drawn as a walk over strata.
We plot the number of active contacts $|A|$, which fixes the stratum's codimension by~\eqref{eq:dim}, against path length, with filled triangles making a contact and open ones breaking one.
Pushing the block past an obstacle cannot be done from one face, so the walk returns to free space and approaches again.
Turning the cube cannot be done from a fixed grasp at all, and the walk becomes a gait.
}
\vspace{-1em}
\label{fig:walk}
\end{figure*}

\section{Planning on the Stratified Configuration Space}
\label{sec:planning}

\looseness=-1
We now plan over the strata of Section~\ref{sec:formalism} without enumerating any of them.
A solution is a \emph{hybrid path} $\sigma$, a walk $A_0 \to A_1 \to \cdots \to A_m$ over strata in which the $k$-th step is a path $\gamma_k$ lying in $\Strat_{A_k}$, with consecutive steps meeting at a configuration where one contact is made or broken (Figure~\ref{fig:strat}(c)).

\looseness=-1
A stratum of positive codimension has measure zero, so we cannot use rejection sampling to sample a configuration from the ambient space.
We therefore sample a set of contacts $A$ together with an ambient configuration, and project that configuration onto the constraint manifold, following the projection-based methods of the constrained planning literature~\cite{jaillet2012path,kingston2018sampling}.
The projection may close the gap of a pair $j \notin A$, and the inequality $\varphi_j \geq 0$ then keeps it from passing through that surface.
The configuration the projection reaches may therefore lie on a stratum other than the one we sampled, and we relabel it by the constraints active there.
A stratum that holds several contacts at once is reached by the same projection, with all of its constraints solved together.

\looseness=-1
We search this stratum graph implicitly with a sampling-based planner, growing a single rapidly-exploring random tree (RRT)~\cite{lavalle2001randomized} whose nodes each store a configuration and the stratum identified by its active set.
Toward a sampled target, the extension attempts three moves in a fixed order.
It first tries to make the contact of the target's active set whose gap at the node is smallest, provided that contact is within reach.
It next tries to move within the current stratum toward the target.
It finally tries to break a contact the target does not hold.
A within-stratum segment is a discrete geodesic, which we trace by taking small steps along $\Strat_A$ toward the target, projecting after each step and evaluating the remaining constraints and the contact cone there.
We measure its length by summing those steps in the ambient metric.

Every edge of the tree is therefore either a segment within one stratum or a transition that makes or breaks one or more contacts.
A path from the root is thus a hybrid path by construction, and we return the first one that reaches the goal.


\section{Preliminary Results}
\label{sec:experiments}

\looseness=-1
We evaluate our approach on two contact-rich manipulation tasks; our implementation uses MuJoCo to evaluate the signed distances and their configuration-space gradients required by the planner.
A Franka robot pushes a T-shaped block around obstacles to an $\mathrm{SE}(2)$ goal, where $q$ holds 7 joint angles and the 3 planar coordinates of the block, so $n = 10$ and $\mathcal{I}$ has 22 body pairs.
A LEAP hand turns a 64\,mm cube by $120^\circ$, where $q$ holds 16 joint angles and the 3 orientation coordinates of the cube, so $n = 19$ and $\mathcal{I}$ has 272 body pairs.
The block coordinates are passive in the first task and the cube orientation is passive in the second, where the fingertip contacts also stick.
We run 48 trials per task and report the success rate, the median planning time, and the median number of contacts made and broken.

\begin{table}[t]
\centering
\caption{
Results over 48 trials per task on a single CPU core.
}
\label{tab:results}
\small
\begin{tabular}{@{}lcccc@{}}
\toprule
 & $n$ & solved & time & make/break \\
\midrule
T-shaped block     & 10 & 48/48 & 760\,ms & 5 / 4 \\
Cube in hand       & 19 & 48/48 & 9.9\,s & 52 / 52 \\
\bottomrule
\end{tabular}
\end{table}

\looseness=-1
Table~\ref{tab:results} reports both tasks, and every trial returns a plan that reaches the goal.
Pushing the block takes a median of 760\,ms and its quickest trial 77\,ms, so the geometry of a ten-dimensional contact problem is cheap enough that the discrete choice needs no search of its own.
The in-hand task takes an order of magnitude longer, and Figure~\ref{fig:walk} shows the reason.
The median block plan makes 5 contacts and breaks 4, while the median in-hand plan makes 52 and breaks 52.
Planning time therefore is consistent with the length of the walk over strata, which is a property of the task rather than a parameter we choose.

\looseness=-1
The in-hand walk of Figure~\ref{fig:walk} is a gait, and the geometry explains why.
A single grasp cannot turn the cube by $120^\circ$, because the fingers reach their joint limits partway through.
The fingertips stick, so holding a contact pins it to one point of the cube, and turning further means breaking that contact, placing the fingertip somewhere new, and turning again.
Repeating that cycle is the only motion the constraints allow, so the gait naturally follows from the stratification rather than from anything we specify in advance.

\vspace{-0.5em}
\section{Limitations and Open Problems}
\label{sec:limitations}

\looseness=-1
We have treated a contact mode as a stratum of the configuration space and a plan as a walk over strata, built from the signed distances and gradients that a collision engine already reports.
A single-tree RRT over these strata solves two contact-rich manipulation tasks within seconds, and the contacts each plan makes and breaks emerge from the search rather than from a specification.
The search returns the first feasible hybrid path, and optimality over such walks remains open.
Our conditions are also kinematic throughout, since the contact cone does not depend on whether an admissible contact wrench can generate the motion.

The consequences of the index itself remain unclear.
Definition~\ref{def:stratified} requires the strata to be connected, and a sticking contact foliates a stratum into leaves indexed by a point of a surface, so two configurations that hold the same contacts can be different states of the search.
The discrete part of our discrete--continuous problem is therefore not finite.
A contact mode is a stratum, but a stratum is a richer object than a mode label, and we have only begun to use the difference.

\bibliographystyle{IEEEtran}
\bibliography{main}

@article{goodwine2002motion,
  title   = {Motion planning for kinematic stratified systems with application to quasi-static legged locomotion and finger gaiting},
  author  = {Goodwine, Bill and Burdick, Joel W},
  journal = {IEEE Transactions on Robotics and Automation},
  volume  = {18},
  number  = {2},
  pages   = {209--222},
  year    = {2002}
}

@phdthesis{goodwine1998thesis,
  author = {Goodwine, Jr., John William},
  title  = {Control of Stratified Systems with Robotic Applications},
  school = {California Institute of Technology},
  year   = {1998}
}

@article{rimon2008general,
  title   = {A general stance stability test based on stratified morse theory with application to quasi-static locomotion planning},
  author  = {Rimon, Elon and Mason, Richard and Burdick, Joel W and Or, Yizhar},
  journal = {IEEE Transactions on Robotics},
  volume  = {24},
  number  = {3},
  pages   = {626--641},
  year    = {2008}
}

@article{han2008convexly,
  title     = {Convexly stratified deformation spaces and efficient path planning for planar closed chains with revolute joints},
  author    = {Han, Li and Rudolph, Lee and Blumenthal, Jonathon and Valodzin, Ihar},
  journal   = {The International Journal of Robotics Research},
  volume    = {27},
  number    = {11-12},
  pages     = {1189--1212},
  year      = {2008},
  publisher = {SAGE Publications}
}

@techreport{bhattacharya2007stratified,
  title       = {Motion Planning in a Stratified Workspace Manifold of a Quadruped Walking Robot},
  author      = {Bhattacharya, Subhrajit},
  institution = {GRASP Laboratory, University of Pennsylvania},
  year        = {2007}
}

@incollection{bretl2005multi,
  title     = {Multi-step motion planning for free-climbing robots},
  author    = {Bretl, Tim and Lall, Sanjay and Latombe, Jean-Claude and Rock, Stephen},
  booktitle = {Algorithmic Foundations of Robotics VI},
  pages     = {59--74},
  year      = {2005},
  publisher = {Springer}
}

@article{jaillet2012path,
  title   = {Path planning under kinematic constraints by rapidly exploring manifolds},
  author  = {Jaillet, L{\'e}onard and Porta, Josep M},
  journal = {IEEE Transactions on Robotics},
  volume  = {29},
  number  = {1},
  pages   = {105--117},
  year    = {2012}
}

@article{kingston2019exploring,
  title     = {Exploring implicit spaces for constrained sampling-based planning},
  author    = {Kingston, Zachary and Moll, Mark and Kavraki, Lydia E},
  journal   = {The International Journal of Robotics Research},
  volume    = {38},
  number    = {10-11},
  pages     = {1151--1178},
  year      = {2019},
  publisher = {SAGE Publications}
}

@article{kingston2018sampling,
  title     = {Sampling-based methods for motion planning with constraints},
  author    = {Kingston, Zachary and Moll, Mark and Kavraki, Lydia E},
  journal   = {Annual Review of Control, Robotics, and Autonomous Systems},
  volume    = {1},
  number    = {1},
  pages     = {159--185},
  year      = {2018},
  publisher = {Annual Reviews}
}

@inproceedings{thomason2024motions,
  title     = {Motions in microseconds via vectorized sampling-based planning},
  author    = {Thomason, Wil and Kingston, Zachary and Kavraki, Lydia E},
  booktitle = {2024 IEEE International Conference on Robotics and Automation (ICRA)},
  pages     = {8749--8756},
  year      = {2024}
}

@inproceedings{sundaralingam2023curobo,
  title     = {Curobo: Parallelized collision-free robot motion generation},
  author    = {Sundaralingam, Balakumar and Hari, Siva Kumar Sastry and Fishman, Adam and Garrett, Caelan and Van Wyk, Karl and Blukis, Valts and Millane, Alexander and Oleynikova, Helen and Handa, Ankur and Ramos, Fabio and others},
  booktitle = {2023 IEEE International Conference on Robotics and Automation (ICRA)},
  pages     = {8112--8119},
  year      = {2023}
}

@article{morozov2025mixed,
  title   = {Mixed discrete and continuous planning using shortest walks in graphs of convex sets},
  author  = {Morozov, Savva and Marcucci, Tobia and Graesdal, Bernhard Paus and Amice, Alexandre and Parrilo, Pablo A and Tedrake, Russ},
  journal = {arXiv preprint arXiv:2507.10878},
  year    = {2025}
}

@inproceedings{toussaint2015lgp,
  author    = {Toussaint, Marc},
  title     = {Logic-geometric programming: an optimization-based approach to combined task and motion planning},
  year      = {2015},
  booktitle = {Proceedings of the 24th International Conference on Artificial Intelligence},
  pages     = {1930--1936}
}

@article{hauser2010multi,
  title     = {Multi-modal motion planning in non-expansive spaces},
  author    = {Hauser, Kris and Latombe, Jean-Claude},
  journal   = {The International Journal of Robotics Research},
  volume    = {29},
  number    = {7},
  pages     = {897--915},
  year      = {2010},
  publisher = {SAGE Publications}
}

@article{englert2020sampling,
  title   = {Sampling-based motion planning on sequenced manifolds},
  author  = {Englert, Peter and Fern{\'a}ndez, Isabel M Rayas and Ramachandran, Ragesh K and Sukhatme, Gaurav S},
  journal = {arXiv preprint arXiv:2006.02027},
  year    = {2020}
}

@incollection{goresky1988stratified,
  title     = {Stratified {Morse} theory},
  author    = {Goresky, Mark and MacPherson, Robert},
  booktitle = {Stratified Morse Theory},
  pages     = {3--22},
  year      = {1988},
  publisher = {Springer}
}

@article{mason1986mechanics,
  title     = {Mechanics and planning of manipulator pushing operations},
  author    = {Mason, Matthew T.},
  journal   = {The International Journal of Robotics Research},
  volume    = {5},
  number    = {3},
  pages     = {53--71},
  year      = {1986},
  publisher = {Sage Publications}
}

@article{lynch1996stable,
  title     = {Stable pushing: Mechanics, controllability, and planning},
  author    = {Lynch, Kevin M. and Mason, Matthew T.},
  journal   = {The International Journal of Robotics Research},
  volume    = {15},
  number    = {6},
  pages     = {533--556},
  year      = {1996},
  publisher = {Sage Publications}
}

@article{goyal1991planar,
  title     = {Planar sliding with dry friction part 1. limit surface and moment function},
  author    = {Goyal, Suresh and Ruina, Andy and Papadopoulos, Jim},
  journal   = {Wear},
  volume    = {143},
  number    = {2},
  pages     = {307--330},
  year      = {1991},
  publisher = {Elsevier}
}

@book{mason2001mechanics,
  title     = {Mechanics of Robotic Manipulation},
  author    = {Mason, Matthew T},
  year      = {2001},
  publisher = {MIT press}
}

@article{chavandafle2020planar,
  title     = {Planar in-hand manipulation via motion cones},
  author    = {Chavan-Dafle, Nikhil and Holladay, Rachel and Rodriguez, Alberto},
  journal   = {The International Journal of Robotics Research},
  volume    = {39},
  number    = {2-3},
  pages     = {163--182},
  year      = {2020},
  publisher = {SAGE Publications}
}

@article{lavalle2001randomized,
  title     = {Randomized kinodynamic planning},
  author    = {LaValle, Steven M. and Kuffner Jr, James J.},
  journal   = {The International Journal of Robotics Research},
  volume    = {20},
  number    = {5},
  pages     = {378--400},
  year      = {2001},
  publisher = {SAGE Publications}
}

\end{document}